\documentclass[10pt,twocolumn,letterpaper]{article}

\usepackage[pagenumbers]{cvpr} % To force page numbers, e.g. for an arXiv version

\usepackage[dvipsnames]{xcolor}

\definecolor{cvprblue}{rgb}{0.21,0.49,0.74}
\usepackage[pagebackref,breaklinks,colorlinks,citecolor=cvprblue]{hyperref}

\usepackage{ifthen}
\usepackage{soul}
\usepackage{multirow}

\usepackage{ifthen}
\definecolor{WeimingColor}{rgb}{0,0,0.8} 
\definecolor{FanColor}{rgb}{0.8,0,0.8}
\newcommand{\weiming}[1]{{\color{WeimingColor} [Weiming: #1]}}
\newcommand{\fan}[1]{{\color{FanColor}[Fan: #1]}}

\newcommand{\delete}[1]{{\color{orange} \st{#1}}}
\newcommand{\final}{1}

\ifthenelse{\equal{\final}{1}}
{
\renewcommand{\weiming}[1]{}
\renewcommand{\fan}[1]{}

\renewcommand{\delete}[1]{}
}
{}

\newcommand{\sysname}{\textit{MotionCrafter}\xspace}

\newcommand{\loss}{\mathcal{L}}
\newcommand{\unet}{\mathcal{U}}
\newcommand{\frozenunet}{\hat{\unet}}
\newcommand{\trainableunet}{\unet_{\theta}}

\def\paperID{} % *** Enter the Paper ID here
\def\confName{CVPR}
\def\confYear{2024}

\title{MotionCrafter: One-Shot Motion Customization of Diffusion Models}
\author{Yuxin Zhang$^{1,2}$\hspace{0.2in} Fan Tang$^{3}$\hspace{0.2in} Nisha Huang$^{1,2}$\hspace{0.2in} Haibin Huang$^4$\\ Chongyang Ma$^4$\hspace{0.2in} Weiming Dong$^{1,2*}$\hspace{0.2in} Changsheng Xu$^{1,2}$\\
$^1$MAIS, Institute of Automation, Chinese Academy of Sciences \hspace{0.2in} $^2$School of AI, UCAS\\
$^3$Institute of Computing Technology, Chinese Academy of Sciences \hspace{0.15in} $^4$Kuaishou Technology \hspace{0.15in} \\
\url{https://zyxelsa.github.io/homepage-motioncrafter/}
}

\begin{document}

\twocolumn[{%
\renewcommand\twocolumn[1][]{#1}%
\maketitle
\begin{center}
    %\centering
    \captionsetup{type=figure}
    \vskip -9mm
    \includegraphics[width=1.0\linewidth]{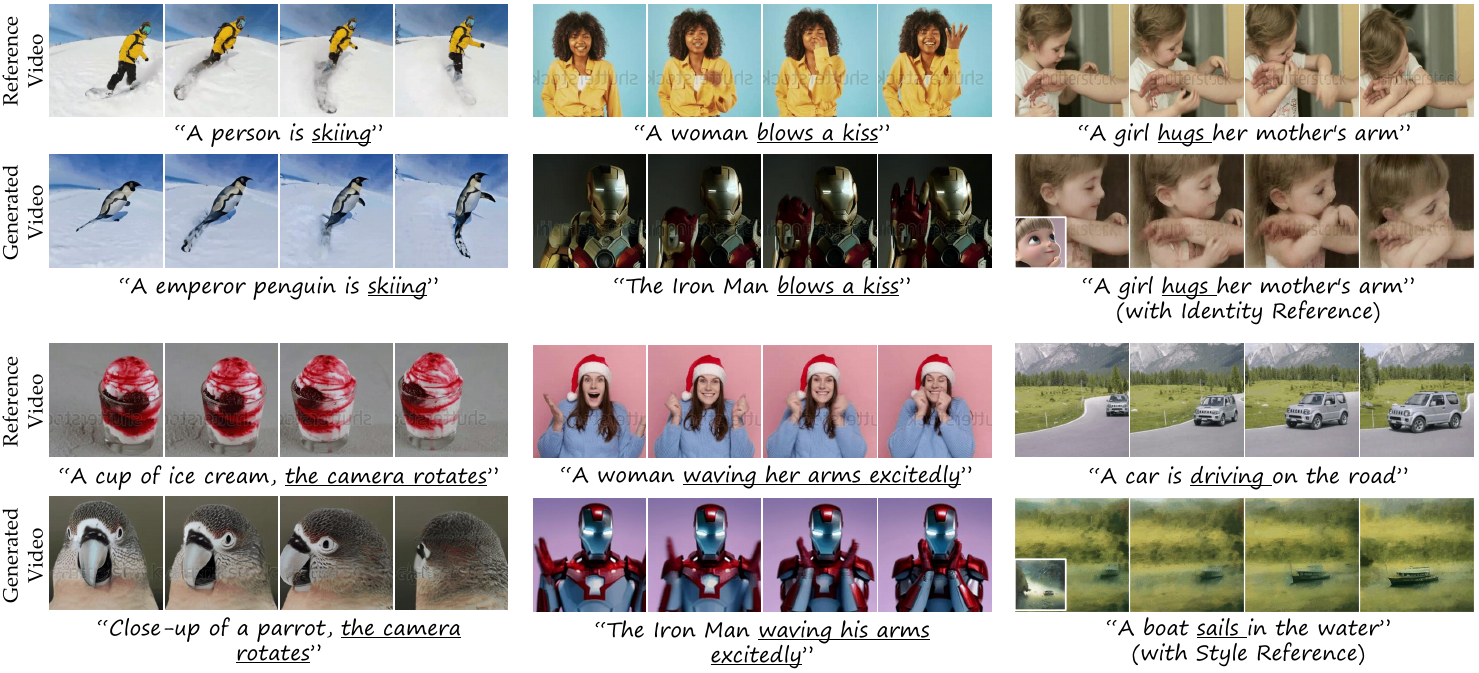}
    \captionof{figure}{
    Motion customization results using \sysname.
    Our method can effectively disentangle the motion features from input video and generate new videos with the same motion but different objects from text prompts (results on the left and in the middle), or with additional image references (results on the right).
    %Given only a single input video, our method can accurately transfer the style attributes such as semantics, material, object shape, brushstrokes and colors of the references  to a natural image with a very simple learned textual description ``[C]''.
    }
    \label{fig:teaser}
    \end{center}%
}]

\begin{abstract}
The essence of a video lies in its dynamic motions, including character actions, object movements, and camera movements. 
While text-to-video generative diffusion models have recently advanced in creating diverse contents, controlling specific motions through text prompts remains a significant challenge.
A primary issue is the coupling of appearance and motion, often leading to overfitting on appearance.
To tackle this challenge, we introduce \sysname, a novel one-shot instance-guided motion customization method.
\sysname employs a parallel spatial-temporal architecture that injects the reference motion into the temporal component of the base model, while the spatial module is independently adjusted for character or style control.
To enhance the disentanglement of motion and appearance, we propose an innovative dual-branch motion disentanglement approach, comprising a motion disentanglement loss and an appearance prior enhancement strategy.
During training, a frozen base model provides appearance normalization, effectively separating appearance from motion and thereby preserving diversity.
% During training, a frozen U-Net provides appearance normalization, effectively separating appearance from motion and thereby preserving diversity.
Comprehensive quantitative and qualitative experiments, along with user preference tests, demonstrate that \sysname can successfully integrate dynamic motions while preserving the coherence and quality of the base model with a wide range of appearance generation capabilities.
Codes are available at \url{https://github.com/zyxElsa/MotionCrafter}.
\end{abstract}

\section{Introduction}
\label{sec:intro}

%% If a picture is worth a thousand words, then a video is worth a million.
Dynamic scenes that resonate with our emotions are not solely memorable due to their captivating visual appeal, but also through the compelling performances of actors, engaging storylines, and meticulous cinematography.
Unlike images, videos encapsulate both spatial and temporal information, resulting in unique dynamic motions.
While current text-to-video techniques can generate videos based on user-input text~\cite{vdm,text2videozero,videofusion}, specific information about actions, object movements, and camera movements in the generated videos often cannot be accurately described by text.
%Thus, it remains a significant challenge to harness the potential of existing pre-trained models to allow users to precisely customize multiple temporal attributes.
Consequently, a significant challenge persists: how can we effectively leverage existing pre-trained models and allow users to precisely customize various temporal aspects in videos. 

% To enhance the controllability and expressiveness of text-to-video generation, we introduce a novel motion customization method.
To enhance controllability and expressiveness, inspired by the customization of text-to-image generation, the concept of motion customization is naturally introduced. 
Motion customization can be considered as a sub-task of video editing and text-to-motion generation, aiming to offer users more precise control over actions, object movements, and camera movements, by allowing them to specify target motions through video inputs.
% within the text-to-video generation framework.
% Our approach allows for the generation of new videos based on elements extracted from a reference video provided by the user.
% Our goal is to enable more precise control over actions, object movements, and camera movements, aligning them better with user expectations through suitable examples.
% The main challenge lies in effectively learning and representing these visual elements within the video, necessitating the separation and control of distinct components within the existing text-to-video generation framework.
% Our method facilitates the creation of new videos by utilizing elements from a user-provided reference video.
% Our goal is to achieve more accurate control over actions, object movements, and camera movements, ensuring they align closely with the user's expectations. 
The primary challenge is to proficiently learn and represent these visual content, requiring the disentanglement and manipulation of temporal elements and network components within existing text-to-video generation framework.
%Our motivation is that by giving suitable examples, the manipulation of actions, object movements, and camera movements can be more accurately controlled to better align with user expectations.
%The primary challenge lies in effectively learning and representing the actions, object movements, and camera movements within the video.
%This necessitates addressing the separation and control of distinct visual elements within the existing text-to-video generation framework.

Several current methods, such as textual inversion (TI)~\cite{gal2023TI}, adopt a personal concept representation technique that maps image references to a text-conditioned space.
This integration with natural language aids in instance-guided concept replication and manipulation.
Other approaches, such as DreamBooth~\cite{ruiz2023dreambooth} and Custom Diffusion~\cite{kumari2023multi}, focus on fine-tuning the core model's internal parameters.
They leverage a set of images provided by the user to augment the model's ability to express specific concepts more effectively.
%Several methods fine-tune the internal parameters of the fundamental model, such as DreamBooth~\cite{ruiz2023dreambooth} and Custom Diffusion~\cite{kumari2023multi}, wherein certain parameters of the Diffusion models are fine-tuned using a user-provided image set to enhance the model's expressive capabilities for specific concepts.
These aforementioned approaches have demonstrated effective performance in tasks involving instance-guided visual customization.
In addition, various video generation methods~\cite{animatediff,he2023animate,gong2023talecrafter} have leveraged them to control artistic styles or specific characters in videos.
However, these image-centric concept representation techniques mainly emphasize appearance, overlooking the essential temporal attributes unique to videos. Consequently, the specific challenge of customizing motion in videos remains unaddressed by these existing approaches.
% However, these image-based concept representation methods primarily focus on appearance and ignore the temporal attributes that are intrinsic to videos.
% As a result, the challenge of motion customization remains unresolved by existing methods.

In this work, we introduce \sysname, a novel one-shot instance-guided method for dynamic motion customization.
To learn customization motion concepts from a given video clip, we fine-tune the pre-trained text-to-video model~\cite{videofusion} and synthesize dynamic renditions in new contexts.
%achieve personalized dynamic motion generation.
Specifically, we introduce a parallel spatial-temporal architecture to infuse motion information into the temporal attention module of the model, aiming at customizing video appearance and motion separately.
To achieve motion disentanglement, we propose a dual-branch motion disentanglement strategy by introducing the base model as a prior.
% which normalizes the fine-tuning of the spatial attention module by introducing a frozen U-Net. 
% Building upon a commonly utilized text-to-video diffusion model~\cite{videofusion}, our method is different in the objective and strategy.
During the training phase, a frozen U-Net is incorporated with the trainable network and both perform inference on the same text conditions, leading to results of the base model and generated output with injected motions, respectively.
Furthermore, we design an appearance prior enhancement scheme where we keep the text prompt for the intended motion fixed, while altering descriptions of various appearances (such as character, scene, etc) during training.
This scheme encourages the frozen base model to generate results with diverse appearances. 
Then, using our proposed motion disentanglement loss, we regulate the mutual information between the generated and reference videos, as well as between the generated results and those from the base model, removing the appearance information from the reference video to achieve motion-appearance decoupling while preserving the generation capability of the base model.

Our contributions can be summarized as follows:
\begin{itemize}
\item %We introduce the instance-guided motion customization that aims at learning motion without the appearance information from given videos for generating new videos.
We present \sysname, a one-shot instance-guided motion customization framework designed to extract appearance-independent motion information from a given video, enabling the generation of new videos.

\item We propose a parallel spatial-temporal architecture that separately fine-tunes the spatial and temporal modules of text-to-video generation models, providing an effective solution for handling both appearance and motion.

\item We propose a dual-branch motion disentanglement approach, which facilitates the separation of motion and appearance by producing diverse appearance priors from the base model.

%\item Experimental results demonstrate that our method can generate a wide range of customized motions, including object movements, human actions, and camera movements. This is achieved without compromising video coherence, thereby achieving state-of-the-art performance.
\item Experiments on a wide range of customized motions, including object movements, human actions, and camera movements, demonstrate the proposed \sysname achieves state-of-the-art performance without compromising video coherence.

\end{itemize}

\section{Related Work}
\label{sec:relatedworks}

\paragraph{Customization of generative models.}
The customization of text-to-image and text-to-video generation models involves the learning of personalized concept with pre-trained models.
Gal \etal~\cite{gal2023TI} propose the task of textual inversion that aims to find a pseudo-word that describes the visual concept of a specific object in a set of user-provided images. 
% Building upon this, Gal \etal~\cite{gal2023encoder} design a word-embedding encoder to predict a new pseudo-word.
Avrahami \etal~\cite{avrahami2023break} introduce a method to extract distinct text tokens for each concept from images containing multiple concepts. 
Besides objects, several methods aim to learn different concepts from given images.
Zhang \etal~\cite{zhang2023inst} propose InST, an attention-based inversion style transfer method.
Huang \etal~\cite{huang2023reversion} propose ReVersion for relation inversion, with the aim of learning specific relations from images.
Wen \etal~\cite{wen2023hard} introduce the concept of hard prompts, which invert a given concept into readable natural languages.
Voynov \etal~\cite{voynov2023pplus} present an extended textual conditioning space consisting of several textual embeddings derived from per-layer prompts, each corresponding to a layer of diffusion model’s denoising U-Net.
Zhang \etal~\cite{zhang2023prospect} reveal that diffusion models generate images by prioritizing low to high frequency information and represent images as a compilation of inverted textual token embeddings generated from per-stage prompts.
Instead of inverting a concept into textual tokens, DreamBooth~\cite{ruiz2023dreambooth} generates a specific subject by finetuning the diffusion models with a unique identifier.
% \cite{StyleDrop} present StyleDrop that enables the synthesis of images in a specific style by adjusting a few trainable parameters in text-to-image models. 
Kumari \etal~\cite{kumari2023multi} propose Custom Diffusion, which optimizes a few parameters in the conditioning mechanism and can be jointly trained for multiple concepts or combine several fine-tuned models.
Chen \etal~\cite{chen2023anydoor} propose AnyDoor, a diffusion-based image generator that can teleport target objects to new scenes at specified locations.

For text-to-video generation, Gong \etal~\cite{gong2023talecrafter} introduce TaleCrafter, an interactive story creation system that can handle multiple characters with layout and structural editing capabilities.
He \etal~\cite{he2023animate} propose a retrieval-based depth-guided method that leverages existing video clips to create a coherent storytelling video by customizing the appearances of characters.
These methods focus on modeling the appearance of visual content, but have difficulty controlling motions.
In contrast, our approach emphasizes the specific inter-frame dynamics of the video.

\paragraph{Text-to-video synthesis and editing.}
Recent studies~\cite{vdm,text2videozero,chen2023videocrafter1} have employed diffusion models to generate lifelike videos, harnessing text as a powerful guiding instruction.
%% To enhance control and quality in the process of video generation, several innovative approaches have been proposed.
VideoFusion~\cite{videofusion} employs a decomposition-diffusion process to enhance control over content and motion in video generation. %By decomposing per-frame noise into fundamental noise shared by all frames and residual noise varying along the time axis, VideoFusion increases flexibility and control.
Furthermore, Imagen Video~\cite{imagenvideo} explores the effectiveness of v-prediction parameterization on sample quality and the progressive distillation-guided diffusion model in video generation.
Similarly, VideoFactory~\cite{videofactory} introduces a novel swapped spatial-temporal cross-attention mechanism that reinforces spatial and temporal interactions. %This approach leverages the intrinsic correlations between spatial and temporal blocks, enhancing coherence and consistency in the generated videos.
% introduces a motion modeling module trained on video clips to distill sensible motion priors. By integrating this module into a frozen text-to-image model, the approach enables the rapid creation of personalized animations.
ModelScopeT2V~\cite{ModelScopeT2V} employs spatial-temporal blocks and a multi-frame training strategy to effectively model temporal dependencies, ensuring smooth motion between frames and improving performance and generalization by learning motion patterns from image-text and video-text datasets.

The above methods care about the consistency between the text and the generated video, while accurately controlling the specific dynamics of the generated video is still challenging.
Several methods aim at controlling videos.
% Gen-1~\cite{gen1} introduces content latent variables based on CLIP embeddings and structural latent variables derived from monocular depth estimation.
ControlVideo~\cite{zhang2023controlvideo} is a training-free framework for text-to-video generation using fully cross-frame interaction in self-attention modules and the generated motions are controlled by edge or depth maps.
Control-A-Video~\cite{chen2023control} incorporates a spatial-temporal self-attention mechanism into the text-to-image diffusion model, enabling video generation based on control signal sequences. %, such as edge or depth maps.
Rerender-A-Video~\cite{rerenderavideo} incorporates hierarchical cross-frame constraints and employs time-aware patch matching and frame blending, maintaining shape, texture, and color consistency in the translated video.
Tune-A-Video~\cite{tuneavideo} introduces a one-shot video tuning method to achieve video editing.
The above methods focus on controlling the structure of the video rather than the dynamic information.
VideoComposer~\cite{wang2023videocomposer} is a textual-spatial-temporal controllable video generation method. They introduce 2D motion vectors that capture pixel-wise movements between adjacent frames, as an explicit control signal to provide guidance on temporal dynamics. 
However, motion vectors are difficult to deal with complex movements and large shape changes.
Make-A-Video~\cite{makeavideo}, Align-your-Latents~\cite{blattmann2023align}, and AnimateDiff~\cite{animatediff} introduce different temporal modules to the latent diffusion model and transform the image generator into a video generator.
They obtain inter-frame consistency by training the temporal module using extensive video data.
Conversely, we inject specific motions into a general temporal module, thereby achieving precise control.

\section{Our Method}
\label{sec:method}
\begin{figure*}
\centering
\includegraphics[width=1\linewidth]{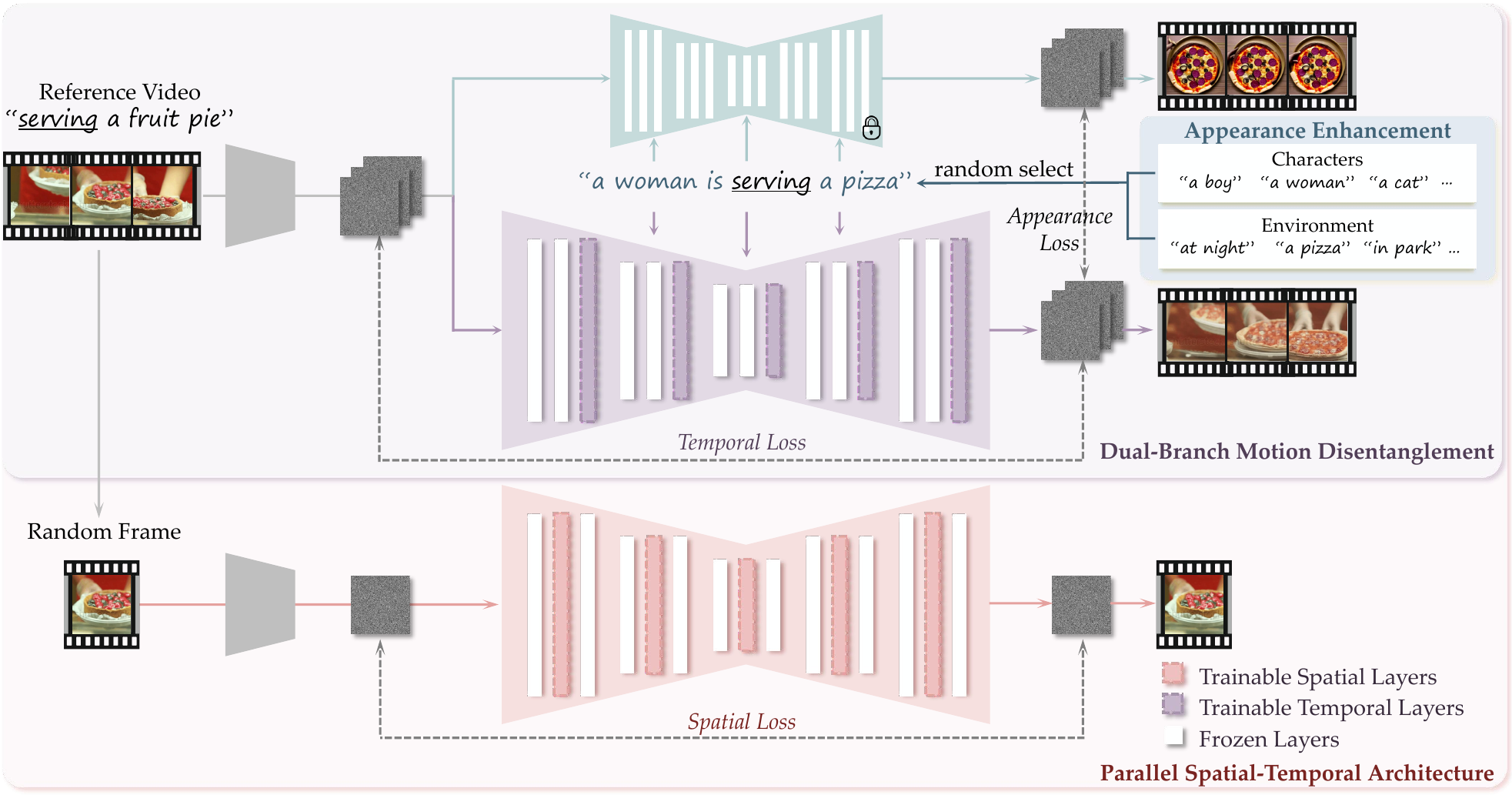}
\caption{The overall pipeline of \sysname.
We use a parallel \textcolor{pink}{spatial}-\textcolor{violet}{temporal} architecture to inject appearance and motion into the corresponding layers in a one-shot fine-tuning manner.
We introduce a \textcolor[rgb]{.64,.7,.68}{frozen U-Net}, which retains the basic model parameters.
During training, \textcolor{violet}{the temporally-tuned U-Net} and \textcolor[rgb]{.64,.7,.68}{the frozen U-Net} receive the same text prompt for \textcolor[rgb]{.25,.45,.5}{appearance enhancement}, resulting in normalization of latents $z_t^{1:N}$ and $\hat{z}_T^{1:N}$.
The appearance normalization loss is used to enforce similarity between the two latents, while $\loss_{temperal}$ is calculated between the generated and reference latents.
}
%\vspace{-2mm}
\label{fig:pipeline}
\end{figure*}

% Pre-trained text-to-video generation diffusion models can represent and produce a wide range of content, but encounter challenges in accurately depicting specific motions.
% The essence of motion lies in the dynamic movement spanning across frames, and this movement is typically coupled with the visual appearance of the subject being presented.
The overall pipeline of \sysname is illustrated in Figure~\ref{fig:pipeline}.
To decompose the appearance and motion of the generated videos, we propose a parallel spatial-temporal architecture (Section~\ref{sec:parallel_architecture}).
It leverages two separate paths to learn the appearance and motion information from videos, corresponding to the spatial and temporal modules in the backbone of a text-to-video generation model.
To achieve better disentanglement, we further design a dual-branch motion disentanglement based on an information bottleneck (Section~\ref{sec:motion_disentanglement}).
We incorporate a frozen branch of the base model to serve as an appearance prior.
During training, the framework takes a reference video and enhanced textual conditioning as inputs and fine-tunes the trainable branch.
% It samples the video at equal intervals and projects them into the latent space with an autoencoder. 
% The dual-branch approach processes the shared latent and obtains two separate latents: one corresponding to the training video information and the other corresponding to the appearance generated by the base model.
During inference, our framework takes user-provided textual conditioning as input and generates results that incorporate the reference video information using only the fine-tuned branch.
% The motion disentangle loss balances the target video and the generated results from the prior preservation, maximizing the preservation of motion information in the video while squeezing out the appearance of the original video.

\subsection{Parallel Spatial-Temporal Architecture}
\label{sec:parallel_architecture}

Our objective is to achieve controllable video generation by harnessing instance-guided motion through powerful pre-trained text-to-video generation models in a cost-effective manner.
To this end, we employ a widely adopted text-to-video diffusion model~\cite{zeroscope} as the foundational architecture for our video generation process. 
This model is structured around a 3D U-Net framework, which operates on the latent space derived from an autoencoder.
At each timestep $t$, the 3D U-Net takes a latent code $z_t$ with dimensions $batch \times frames \times width \times height \times channels$ as input, and gives the predicted noise $\epsilon_\theta(\cdot)$. 
Additionally, textual conditions $\tau_\theta(y)$ are incorporated to provide contextual guidance.
The U-Net backbone consists of down blocks, mid blocks, and up blocks, each accompanied by spatial and temporal attention modules. The spatial attention module performs operations on the 2D spatial dimensions encompassing the width and height of the latent codes. The temporal attention module captures inter-frame relationships along the frame dimension.
The inherent architecture of text-to-video diffusion models naturally segregates spatial and temporal structures. 

In our approach, we propose a parallel spatial-temporal learning framework that leverages the intrinsic properties of the temporal and spatial attention modules within the text-to-video diffusion models.
Our framework leverages separate fine-tuning of these modules to learn appearance and motion features from a reference video.
Specifically, for the appearance information, while leaving other modules unchanged, we randomly sample a frame from the reference video and fine-tune the spatial attention module. This fine-tuning process is guided by the spatial loss, which is defined as follows:
\begin{equation}
\begin{aligned}
% \resizebox{.9\hsize}{!}{\loss_{spatial}=\mathbb{E}_{\mathcal{E}\left(x_0^{i}\right), y, \epsilon \sim \mathcal{N}(0, I), t,i \sim \mathcal{U}(0, N)}\left[\left\|\epsilon-\epsilon_\theta\left(z_t^i, t, \tau_\theta(y)\right)\right\|_2^2\right],}
&\mathcal{L}_{spatial}=\\
&\mathbb{E}_{\mathcal{E}\left(x_0^{i}\right), y, \epsilon \sim \mathcal{N}(0, I), t,i \sim \mathcal{U}(0, N)}\left[\left\|\epsilon-\epsilon_\theta\left(z_t^i, t, \tau_\theta(y)\right)\right\|_2^2\right],
\end{aligned}
\end{equation}
where $\epsilon$ is the added noise, $N$ is the number of frames of the reference video. 
$x_0^{i}$ is the sampled frame of the reference video $x_0^{1: N}$, and $z_t^i$ is the sampled frame from the latent code $z_t^{1: N}$.
For the motion information, we fine-tune the temporal modules in the down blocks, mid blocks, and up blocks of the 3D U-Net to update the correlations along the temporal dimension.
The temporal loss is formulated as:
\begin{equation}
\begin{aligned}
&\mathcal{L}_{temporal}=\\
&\mathbb{E}_{\mathcal{E}\left(x_0^{1: N}\right), y, \epsilon \sim \mathcal{N}(0, I), t}\left[\left\|\epsilon-\epsilon_\theta\left(z_t^{1: N}, t, \tau_\theta(y)\right)\right\|_2^2\right].
% \mathcal{L}_{temporal}=\mathbb{E}_{z_0, y, \epsilon \sim \mathcal{N}(0, I), t \sim \mathcal{U}(0, T)}\left[\left\|\epsilon-\epsilon_\theta\left(z_t, t, \tau_\theta(y)\right)\right\|_2^2\right],
\end{aligned}
\end{equation}
However, during the training and inference stages, the information represented by the spatial and temporal modules is coupled together.
Therefore, the challenge of customizing motions based on pre-trained text-to-video generation models lies in decomposing the spatial and temporal information of the generated video.

\subsection{Dual-Branch Motion Disentanglement}
\label{sec:motion_disentanglement}

\begin{figure}
\centering
\includegraphics[width=1\linewidth]{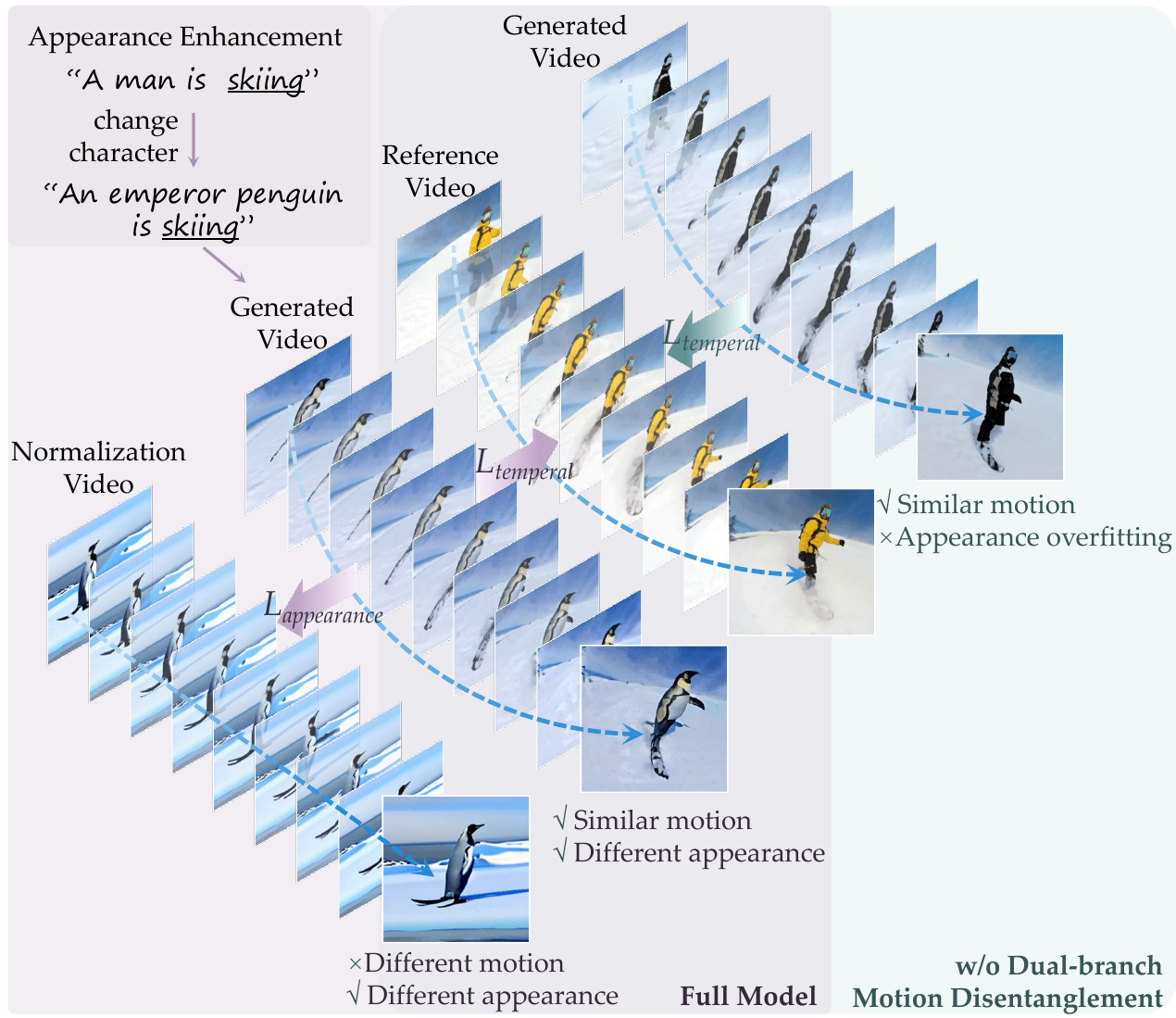}
\caption{
The workflow of Dual-branch Motion Disentanglement enables the model to strike a balance between the reference video and the normalization video. 
Thus, \sysname can maintain the base model's appearance while have a consistent motion as the reference video. 
Without the motion disentanglement, as illustrated on the right, one-shot customization could lead to overfitting of the appearance to the reference videos.}
% \vspace{-3mm}
\label{fig:disentanglement}
\end{figure}

\paragraph{Motion disentanglement loss.}
To address the aforementioned coupling issue of spatial and temporal information, we introduce a dual-branch framework for motion disentanglement in videos, as illustrated in Figures~\ref{fig:pipeline} and~\ref{fig:disentanglement}.
The fitting of the model to the appearance of input videos leads to the inherent loss of its own diversity. By introducing a base model as a prior, the diversity can be better preserved, thereby alleviate the issue of appearance overfitting. 
During the training process, we incorporate an additional frozen U-Net maintaining the parameters of the base model to provide normalization videos.
To separate appearance information from the reference video, we introduce a motion disentanglement loss based on the information bottleneck, which consists of an appearance normalization loss that pushes the generated results to match the normalization videos, and the aforementioned temporal loss encouraging the model to generate results consistent with the reference video.
Thus, by controlling the accessibility of information bottlenecks, we can effectively eliminate appearance information from the reference video while avoiding overfitting. 
This approach ensures the preservation of the pre-trained model's appearance diversity, leading to improved controllability of the generated videos.

Specifically, in the autoencoder's latent space, the dual-branch U-Net consists of a frozen U-Net backbone $\frozenunet$ with the original weights and another U-Net $\trainableunet$ with trainable temporal layers.
At each timestep, a shared latent code undergoes separate processing by the two branches, resulting in $\hat{z}_{t}$ and $z_{t}$. In this process, $\hat{z}_t$ preserves the diverse appearance generated by the frozen model, while the reference information is injected into $z_{t}$ via the trainable branch. 
We propose an appearance normalization loss $\loss_{appearance}$, which imposes a constraint on the KL divergence between the distributions of the latent codes $z_{t}$ and $\hat{z}_{t}$.
By aligning the distributions, the appearance information of the reference video is squeezed out. 
This process is parameterized as:
\begin{equation}
\begin{aligned}
\loss_{appearance}= D_{K L}\left(q_{\theta}(z_t \mid x_t) \| p(\hat{z}_t)\right).
 % \mathcal{L}_{motion}=&\mathbb{E}_{q_\phi(\mathbf{z} \mid \mathbf{x})}\left[\log p_\theta(\mathbf{x} \mid \mathbf{z})\right] \\
 % &+\beta D_{K L}\left(q_\phi(\mathbf{z} \mid \mathbf{x}) \| p(\mathbf{z})\right),
\end{aligned}
\end{equation}
The appearance normalization loss described above is combined with a temporal loss $\loss_{temperal}$, thereby facilitating the extraction of motion.
The full objective is named as our motion disentanglement loss $\loss_{motion}$ and is formulated as:
\begin{equation}
\begin{aligned}
\loss_{motion}=\loss_{temperal}+\beta\loss_{appearance},
\end{aligned}
\end{equation}
where the hyper-parameter $\beta$ is included to control the accessibility of information bottlenecks.
We set $\beta=5$ in all the experiments.

\paragraph{Appearance prior enhancement scheme.}
To better preserve the diversity of appearance in the generated backbone, we propose an appearance prior enhancement scheme.
This scheme is designed to encourage the original U-Net model to generate more diverse content, while simultaneously preserving the intended motion.
Specifically, we create templates that capture a variety of appearance, including descriptions of diverse objects and scenes in natural language (\eg, ``a woman with a hat is \{\} in a park'').
By incorporating the appearance details from these templates and combining them with the target motion (\ie, replacing the \{\} with a description of the motion), the original U-Net $\frozenunet$, which remains frozen, is able to generate results with more diverse appearance while maintaining the motion information.
This scheme helps the network differentiate between appearance and motion information, leading to a more comprehensive disentanglement.
   
\section{Experiments}
\label{sec:experiments}

\begin{table*}%[t]
\centering
\caption{Quantitative evaluation and user study results.
The best numbers are in \textbf{bold} and the second best results are \underline{underlined}.}
\vskip -3mm
\resizebox{\linewidth}{!}{
\begin{tabular}{c||c|c||c|c|c|c}
\toprule
\multirow{2}{*}{Method}&\multicolumn{2}{c||}{Quantitative evaluation}&\multicolumn{4}{c}{User preference}\\
\cline{2-3} \cline{4-7}
&Diversity$\uparrow$&Consistency$\uparrow$&Motion fidelity$\uparrow$&Diversity$\uparrow$&Consistency$\uparrow$&Visual quality$\uparrow$\\
% \hline
\midrule
Zeroscope~\cite{zeroscope}&\underline{0.2475}&0.8941&2.05&2.57&2.70&2.46\\
\hline 
Control-A-Video~\cite{chen2023control}&0.2114&0.8019&\underline{3.22}&2.48&2.94&2.54\\
\hline 
ControlVideo~\cite{zhang2023controlvideo}&0.2272&0.8691&1.65&2.36&2.08&2.03\\
\hline 
VideoComposer (depth)~\cite{wang2023videocomposer}&0.2423&0.8952&2.95&2.88&\underline{3.08}&\underline{2.73}\\
\hline
VideoComposer (motion)~\cite{wang2023videocomposer}&0.2474&\textbf{0.8984}&2.70&\underline{2.90}&2.98&2.71\\
\hline
Rerender-A-Video~\cite{rerenderavideo}&0.1829&0.8355&3.07&2.01&2.46&2.27\\
\hline
Tune-A-Video~\cite{tuneavideo}&0.2184&0.8097&2.70&2.49&2.32&2.27\\
\midrule
Ours&\textbf{0.2559}&\underline{0.8975}&\textbf{4.09}&\textbf{4.23}&\textbf{4.12}&\textbf{3.98}\\
%\hline
\bottomrule
\end{tabular}
}
%\vspace{-3mm}
\label{tab:clip_evaluation}
\end{table*}

\begin{figure*}
\centering
\includegraphics[width=\linewidth]{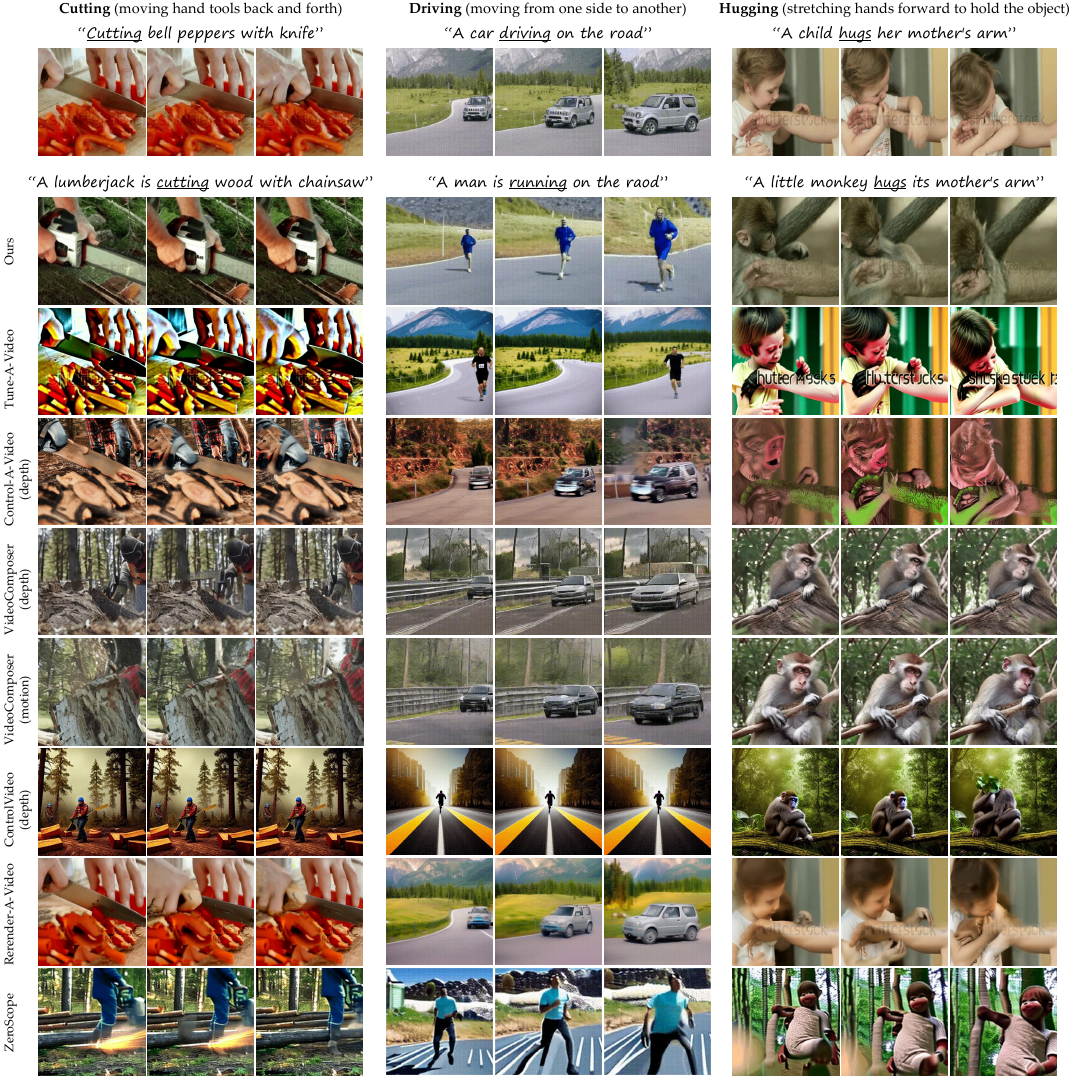}
\caption{
Qualitative evaluation results. Our method outperforms state-of-the-art methods in appearance diversity and motion fidelity.}
% \vspace{-3mm}
\label{fig:comparison}
\end{figure*}

\begin{figure*}
\centering
\includegraphics[width=1\linewidth]{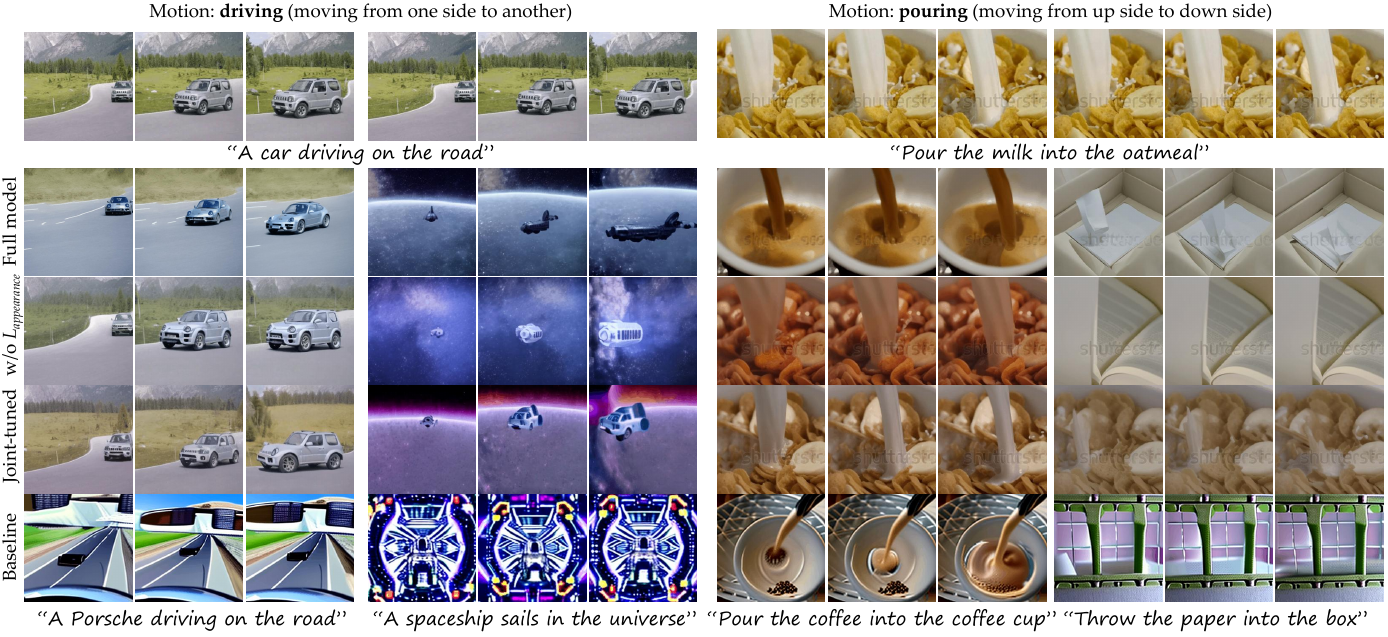}
\caption{
Ablation study results of our parallel spatial-temporal architecture and dual-branch motion disentanglement.}
%\vspace{-3mm}
\label{fig:ablation}
\end{figure*}

In this section, we demonstrate that \sysname is capable of replicating motions from the reference video while maintaining content coherence. Additionally, it offers greater editability compared to state-of-the-art text-to-video customization baselines.

\subsection{Experimental Setup}

\paragraph{Methods for comparison.}
We compare our approach with state-of-the-art text-to-video generation method ZeroScope~\cite{zeroscope}, several video-to-video editing methods including Control-A-Video~\cite{chen2023control}, VideoComposer~\cite{wang2023videocomposer}, Rerender-A-Video~\cite{rerenderavideo}, ControlVideo~\cite{zhang2023controlvideo}, as well as the fine-tuning-based video customization method Tune-A-Video~\cite{tuneavideo}.

\paragraph{Evaluation dataset.}
To ensure a fair comparison, we utilize widely used video segments from previous papers, along with clips from the WebVid-10M dataset~\cite{bain2021frozen}.
We carefully select 12 motions for qualitative and quantitative evaluations.
These motions include serving an item, pouring, cutting, hugging, taking out, shaking, counting money, back flipping, taking a selfie, giving an item, driving, and running.
For each motion, we employ three basic prompts that alter either the object or background, alongside three more complex prompts that involve changes to both the object and background.
Consequently, we obtain a total of 72 video clips for each method to compare.
%We selected 12 mitions for qualitative and quantitative evaluations, including serving an item, pouring, cutting, hugging, taking out, shaking, counting money, backflipping, taking a selfie, giving an item, driving, and running. For each motion, we used three easy prompts (changing only the object or background) and three hard prompts (changing both the object and background, etc.). In total, we obtained 72 video clips for each method.

% \paragraph{Implementation details.}
% In all of our experiments, we use ZeroScope~\cite{zeroscope} with the default network architecture.
% % The input video resolution is $256\times 256$ and is sampled into 16 frames.
% The resolution of the input video is $256 \times 256$ and each sequence contains $16$ frames.
% % The training process for each motion requires approximately $150 \sim 300$ iterations using an NVIDIA L40 with a batch size of 1.
% The number of inference steps is set to $T = 25$ and the guidance scale is set to $w = 7.5$.

\subsection{Quantitative Evaluations}

We measure the appearance diversity using the average CLIP~\cite{clip} similarity between the diverse text prompts and all frames of the generated videos. We measure the temporal consistency using the average CLIP similarity between adjacent frames. Table~\ref{tab:clip_evaluation} presents the quantitative evaluation results of our method and the seven baseline approaches. We achieve state-of-the-art results in both metrics.
%% VideoComposer~\cite{wang2023videocomposer} achieves good diversity and consistency results but may not faithfully transfer the target motion, as shown in the user preference and qualitative comparison results.
VideoComposer~\cite{wang2023videocomposer} delivers desirable results in terms of diversity and consistency. However, as indicated by our user study results and qualitative comparisons, it may fall short in accurately transferring the target motion.

\subsection{Qualitative Evaluations}

As shown in Figure \ref{fig:comparison}, we conduct qualitative comparisons with seven state-of-the-art methods. To highlight \sysname's robust ability to decouple motion and appearance, we employ complex prompts that involve substantial content changes between the reference and generated videos, such as ``knives'' transforming into ``chainsaws''.

ZeroScope~\cite{zeroscope} serves as our baseline model and is capable of generating videos with the desired appearance. Due to the lack of motion control, it struggles to produce results that corresponded to the target actions.
Tune-A-Video~\cite{tuneavideo} is a fine-tuning-based method similar to ours, but its objective does not focus on decoupling motions. Therefore, it fails to handle significant changes in image appearance and tends to generate results similar to the original video.
We employ the depth-map control model of Control-A-Video~\cite{chen2023control} to minimize the impact of the original video's appearance. However, it fails to alter the shape of objects, generating a car in the second example and a distorted monkey in the third example.
We use VideoComposer~\cite{wang2023videocomposer} with both depth-map control and motion control models. The depth-map control also faces challenges in altering the shapes of objects. VideoComposer represents video-specific elements using motion vectors, \ie, 2D vectors that capture pixel-wise movements between adjacent frames. However, this motion representation method fails to capture fine-grained motion patterns, such as the arm movement in the third example.
Rerender-A-Video~\cite{rerenderavideo} supports complex control conditions and can effectively preserve the content of the input video. However, it encounters difficulties when the prompt significantly differs from the content in the reference video.
We utilize the depth-map control model of ControlVideo~\cite{zhang2023controlvideo}. The depth-wise constraint is relatively less restrictive, allowing for drastic changes in the appearance to match the specified style. However, this method results in the loss of motion information from the reference video.
\sysname generates desirable content while ensuring consistent actions and high visual quality.

\subsection{User Study}

We conduct a user preference assessment, comparing our approach with the aforementioned seven methods.
We employ a rating scale with four criteria: motion fidelity, appearance diversity, video consistency, and visual quality, applied to a dataset of 12 actions.
In total, 102 participants took part in the survey.
They were first informed about the objectives and settings of the motion customization task. Subsequently, we showed them reference videos, outputs from eight different methods, and the corresponding prompt conditions.
The participants were asked to rate the outputs of each method using a five-point scale, with higher scores reflecting increased user satisfaction with the generated results.
The user study results are presented in Table~\ref{tab:clip_evaluation}. 
Our method achieves the highest user preference, particularly in terms of motion fidelity and appearance diversity.

\subsection{Ablation Study}
We conduct ablation study to validate the effectiveness of the two key components of our method, \ie, parallel spatial-temporal architecture and dual-branch motion disentanglement.
The ablation study results are presented in Figure~\ref{fig:ablation}, along with a comparison to the baseline model ZeroScope~\cite{zeroscope}.
In the third row, it is noticeable that, even without employing $\loss_{appearance}$ to separate appearance from motion, the model successfully generates diverse content but still retains the appearance features of the reference video.
For instance, in the first example, the shape of the vehicle lies between the reference video and the target style. In the third example, the coffee undergoes a color change but maintains the texture of the cereal.
Additionally, the model's ability to learn motion is weakened, as demonstrated in the fourth example, where the dynamic effect is lost in the output.
In the fourth row, it can be observed that when fine-tuning the temporal and spatial modules of the text-to-video diffusion model with the reference video, the model learns coupled information.
For instance, in the second example, the spaceship still maintains the style of the vehicle in the reference video, giving the impression of a car traveling in space.
In both the third and fourth examples, the appearance of the generated videos remains relatively unchanged.
The bottom row shows the corresponding results of ZeroScope~\cite{zeroscope}. It can be observed that for most simple prompts, the model is able to generate reasonable outputs. However, for some challenging prompts, such as ``Throw the paper into the box'', the model struggles to generate desirable results.
Compared to these alternative baselines, our full model produces superior results, particularly in motion fidelity and appearance diversity.

\section{Conclusion}
\label{sec:conlusion}

In this work, we tackle the challenge of motion customization in text-to-video generation by proposing a one-shot instance-guided approach. 
The proposed parallel spatial-temporal architecture can effectively separate motion and appearance, enabling the injection of reference motions into the temporal module of the base model.
Additionally, our novel dual-branch motion disentanglement method successfully decouples appearance and motion, by incorporating a motion disentanglement loss and an appearance prior enhancement scheme.
%% By leveraging a frozen U-Net and carefully balancing the influence of reference videos and appearance normalization, our method achieves impressive results in generating videos with diverse appearances and specific dynamic motions.
The extensive quantitative and qualitative evaluations, as well as the user preference survey, demonstrate the effectiveness of \sysname. 
Our work paves a way for more motion-aware video generation.
%% Exploring techniques to improve the model's capability to generate high-quality group objects would also be an important direction for further research.

% \subsection{Limitations}
\paragraph{Limitations and future work.}
\label{sec:limitation}

While we have demonstrated the ability of \sysname to generate complex dynamic motions, there are certain limitations imposed by the model structure and computational resources. 
% For complex actions that require a longer duration to complete, such as a set of aerobics exercises, even with the inclusion of stride sampling, they often require more than 24 frames to learn in order to maintain action coherence.
In cases of complex actions requiring extended durations for completion, such as a series of aerobics exercises, maintaining action coherence often necessitates learning from more than 24 frames. %, even when stride sampling is included.
% To address this issue, one possible solution is to divide the sequential actions into several units and learn separately, or introduce interpolation between frames.
One potential approach to tackle this issue is by segmenting the sequential actions into multiple units, or by implementing interpolation between frames.
Furthermore, for complex actions involving a group of individuals, such as scenes from a ballet group performance, \sysname may struggle to accurately capture the detailed dynamics of each individual.
% This is partly due to the complexity of these dynamic motions and also because current text-to-video generation models do not possess the capability to generate high-quality group objects.
This challenge arises partly from the intrinsic complexity of motions and is also due to the limitations of current text-to-video generation models in producing high-quality representations of group objects.
Addressing the aforementioned challenges will be targeted in our future work.
% In future work, we plan to focus on addressing the limitations of our method, such as the inability to handle complex and long-duration motions or group actions involving multiple instances. 

{
    \small
    \bibliographystyle{ieeenat_fullname}
    \bibliography{ShotCrafter}
}


\end{document}

\clearpage
\setcounter{page}{1}
\twocolumn[{%
\renewcommand\twocolumn[1][]{#1}%
\maketitlesupplementary
\begin{center}
    \captionsetup{type=figure}
    % \vskip -9mm
    \includegraphics[width=0.95\linewidth]{images/animate}
    \captionof{figure}{
    Results of the integration of our \sysname with AnimateDiff~\cite{animatediff}. Our method demonstrates strong generalization performance and can generate high-quality animation results with controllable motions within the AnimateDiff framework.}
    \label{fig:animate}
    \end{center}%
}
]

\input{sec/X_suppl}

%\newpage
{
    \small
    \bibliographystyle{ieeenat_fullname}
    \bibliography{ShotCrafter}
}
